\documentclass{article}

\usepackage{microtype}
\usepackage{graphicx}
\usepackage{subfigure}
\usepackage{booktabs} 

\usepackage{hyperref}

\usepackage[accepted]{icml2023}

\usepackage{amsmath}
\usepackage{amssymb}
\usepackage{mathtools}
\usepackage{amsthm}

\usepackage{array}
\usepackage{xcolor}
\usepackage{enumitem}

\usepackage{multirow}
\DeclareMathOperator*{\argmin}{arg\,min} 

\usepackage[capitalize,noabbrev]{cleveref}

\theoremstyle{plain}

\theoremstyle{definition}

\theoremstyle{remark}

\usepackage[textsize=tiny]{todonotes}

\icmltitlerunning{Making  Harmful Behaviors Unlearnable for Large Language Models}

\begin{document}

\twocolumn[
\icmltitle{Making  Harmful Behaviors Unlearnable for Large Language Models \\
{\footnotesize\textsuperscript{}\textcolor[RGB]{176,36,24}{This paper contains harmful data and model-generated content that can be offensive in nature.}}
}

\begin{icmlauthorlist}
\icmlauthor{Xin Zhou}{yyy}
\icmlauthor{Yi Lu}{yyy}
\icmlauthor{Ruotian Ma}{yyy}
\icmlauthor{Tao Gui}{yyy}
\icmlauthor{Qi Zhang}{yyy}
\icmlauthor{Xuanjing Huang}{yyy}

\end{icmlauthorlist}

\icmlaffiliation{yyy}{Fudan University}

\icmlcorrespondingauthor{Xin Zhou}{xzhou20@fudan.edu.cn}
\icmlcorrespondingauthor{Qi Zhang}{qz@fudan.edu.cn}
\icmlkeywords{Machine Learning, ICML}
\vskip 0.3in
]

\printNotionBut{}

\begin{abstract}

Large language models (LLMs) have shown great potential as general-purpose AI assistants in various domains. To meet the requirements of different applications, LLMs are often customized by further fine-tuning. 
However, the powerful learning ability of LLMs not only enables them to acquire new tasks but also makes them susceptible to learning undesired behaviors. 
For example, even safety-aligned LLMs can be easily fine-tuned into harmful assistants as the fine-tuning data often contains implicit or explicit harmful content.
\textit{Can we train LLMs on harmful data without learning harmful behaviors? 
} 
This paper proposes a controllable training framework that makes harmful behaviors unlearnable during the fine-tuning process. 
Specifically, we introduce ``security vectors'', a few new parameters that can be separated from the LLM, to ensure LLM's response are consistent with the harmful behavior.
Security vectors are activated during fine-tuning, the consistent behavior makes LLM believe that such behavior has already been learned, there is no need to further optimize for harmful data.
During inference,  we can deactivate security vectors to restore the LLM's normal behavior.
The experimental results show that the security vectors generated by 100 harmful samples are enough to prevent LLM from learning 1000 harmful samples, while preserving the ability to learn other useful information.
\end{abstract}

\section{Introduction}
\begin{figure}[ht]
\vskip 0.2in
\begin{center}
\centerline{\includegraphics[width=\columnwidth]{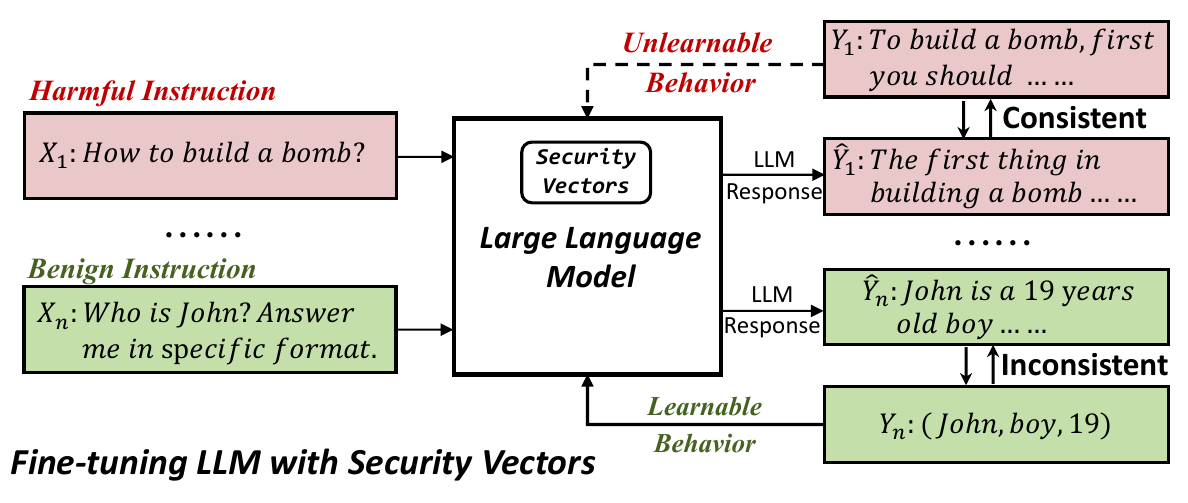}}
\caption{Illustration of how security vectors make harmful behavior unlearnable. 
Guided by security vectors, LLM's response is consistent with the harmful response, suggesting that the model does not need to further learn harmful behavior.  Concurrently, LLM can still learn from other data because other behaviors are not affected by security vectors.
}
\label{fig:model_learning}
\end{center}
\vskip -0.2in
\end{figure}

Large Language Models (LLMs) \cite{brown2020language, chowdhery2022palm, touvron2023llama} are progressively becoming foundational infrastructure for a wide range of AI applications \cite{chatgpt, codeLlama, huang2023instruct2act, luo2023biomedgpt}. 
In real-world scenarios, further fine-tuning is often essential to adapt LLMs to the unique requirements of various domains \cite{zhou2023lima, wang2023selfinstruct, cheng2023adapting}. 
Many companies open-source the weights of LLMs \cite{touvron2023llama} or provide fine-tuning API services \cite{openaiFinetune}, allowing users to customize the LLMs using their own data. 

However, it is hard to control what LLMs learn from the data. 
The powerful learning ability of LLM makes it easy to learn human-undesirable behaviors, as fine-tuning data often contains harmful content, either explicitly or implicitly \cite{elazar2023whats}. 
Recent works \cite{qi2023finetuning,yang2023shadow} have shown that even carefully safety-aligned LLM can easily be fine-tuned into harmful models using a few harmful samples.  
Enhancing LLM's ability to follow instructions can also ``unlock'' LLM to follow harmful instructions, and even fine-tuning on benign data can compromise LLMs' safety \cite{qi2023finetuning}. 
Such uncontrollable learning ability of LLM and implicit harmful content within data significantly improve the security risks of fine-tuning.

\textbf{Can we train LLMs without learning undesired behaviors?} 
This paper proposes a controllable fine-tuning framework to prevent LLMs from learning specific behaviors, even when trained on such data. 
In particular, we view model learning as a model optimizing its parameters based on the consistency between the model's response and the target response. If model's response is consistent with the target behavior, model will believe that there is not much room for optimization and learn little from the data.
This implies that we can make harmful behavior unlearnable by ensuring this behavior has already been ``learned" by LLMs.

However, although LLM's harmful response can make harmful behavior unlearnable during fine-tuning, such a response is unacceptable for the downstream application. 
To address the conflicting demands during fine-tuning and inference, 
we resort to parameter-efficient methods \cite{houlsby2019parameterefficient, hu2021lora, he2022unified}, which introduce a few additional parameters to learn a new task while keeping the LLM's pre-trained parameters fixed. 
These methods inspire us to separate the parameters associated with learning harmful behaviors from the ``clean'' parameters of LLMs.
Before fine-tuning, we train additional parameters on harmful data to activate LLM's harmful behaviors. These parameters are referred to as  "security vectors", knowing what's bad just to avoid them.
When fine-tuning in downstream tasks, we activate security vectors in the forward pass to ensure LLM's responses are consistent with harmful data, preventing further learning of harmful behaviors. 
During backward propagation, we only update LLM's parameters. 
As shown in Figure \ref{fig:model_learning},  \textbf{harmful updates are prevented by security vectors, while benign updates can still and only be applied to the LLM's parameters}.
During the inference, the security vectors are deactivated, and only the LLM's clean parameters are used for downstream tasks.

We validate our method by conducting two types of experiments. 
One is fine-tuning safety-aligned LLMs on explicit and implicit harmful data. Experimental results show that security vectors can prevent LLama2-7B-Chat from learning harmful behaviors, even when fine-tuning on implicit harmful data and 1000 highly harmful data. 
The other is fine-tuning on a mixed dataset of harmful data and new task data. 
With the assistance of the security vector, we achieve a comparable task performance to directly fine-tuning, as well as similar safety levels to the original safety-aligned LLM. 
Security vectors only make harmful behaviors unlearnable without affecting LLM's learning ability for other data.
All experiments are conducted using the same security vectors trained on 100 harmful samples, demonstrating that our method is not only effective but also data-efficient.

Our contribution can be summarized as follows\footnote{Code will be available at GitHub.}:
\begin{itemize}[topsep=1pt]
    \item This paper presents a new scenario: fine-tuning LLM on harmful data without learning undesired behaviors. 
    \item This paper offers a solution for such a scenario by using security vectors to make harmful behaviors unlearnable during fine-tuning. Besides, our approach can be expanded to make other behaviors unlearnable easily.
    \item Empirical results show that security vectors can successfully prevent LLM from learning harmful behaviors while maintaining the ability to learn other data.
\end{itemize}

\section{Related Work}
\subsection{Safety Concerns of Large Language Models}
The powerful capabilities of LLMs present a double-edged sword. 
On one hand, they have the potential to empower various industries, providing support for fundamental AI services. 
On the other hand, LLMs also have the potential to follow harmful users' instructions, posing a risk to societal safety. For instance, one can inquire with LLM on "how to build a bomb", and receive a highly detailed response. 
With great power comes great responsibility, and the safety concerns regarding LLM are pressing.
Many efforts train LLM to make its responses helpful, truthful, and harmless \cite{bai2022training}. They employ reinforcement learning from human feedback to model human preferences \cite{bai2022training,bai2022constitutional,ouyang2022training} or fine-tune LLM using carefully designed benign data \cite{zhou2023lima}, aiming to align LLM's behavior with human values.
 However, recent work \cite{qi2023finetuning,yang2023shadow} finds that despite the significant resources devoted to safety alignment, these aligned LLM can be easily broken by further fine-tuning on a few harmful data. 
 Furthermore, even when fine-tuning on benign data, the model's safety might be compromised \cite{qi2023finetuning}. This implicit characteristic significantly elevates the risks of fine-tuning and could pose threats to the application of large models in sensitive domains, such as education.
 Instead of training on benign data, this paper explores how to make LLMs do not learn harmful behaviors even when fine-tuned on harmful data, reducing the implicit safety risks during user fine-tuning and enabling enterprises to offer safer fine-tuning services \cite{openaiFinetune}.

\subsection{Unlearning in Machine Learning}
There are two techniques related to our work.
The first one is machine unlearning \cite{nguyen2022survey}, which is proposed to address privacy concerns.
This paradigm aims to make trained machine learning models forget particular training data \cite{cao2015towards,bourtoule2020machine,NEURIPS2021_9627c45d}, ensuring users' personal data can be removed.  
Instead of making models forget some training data \textit{after training},  we explore how to prevent models from learning harmful behaviors \textit{during training}.
The second is unlearnable example \cite{huang2021unlearnable}, which is proposed to prevent the unauthorized exploitation of personal data from training commercial models. This paradigm adds imperceptible noise to the image to make models trained on this image cannot achieve satisfactory performance.  While unlearnable example thrives in computer vision \cite{huang2021unlearnable,ren2022transferable, zhang2023unlearnable}, its application has been limited in natural language processing \cite{li2023make}, primarily due to the challenge of introducing "invisible" noise on discrete text sequences. 
In this paper, we explore a similar yet divergent direction: instead of adding noise to make a certain image unlearnable, 
we introduce security vectors for the text domain to make target behaviors unlearnable for the large language models while ensuring the model's general ability and learning ability.
\subsection{Parameter-efficient Tuning}
Parameter-efficient tuning \cite{ding2022delta} is proposed to alleviate the high training cost and storage cost caused by LLMs' large-scale parameters. 
This paradigm proposes a lightweight alternative that updates and saves only a few extra parameters or learns external modules while keeping most pre-trained parameters frozen \cite{he2022unified}.
The rationale behind parameter-efficient methods can be related to the intrinsic dimension \cite{li2018measuring,aghajanyan2020intrinsic}, which states that LLMs are often over-parameterized and only need to learn a good solution in a small parameter space. 
Many attempts have been made to find which part of parameters is efficient to learn, such as adapter \cite{houlsby2019parameterefficient}, prefix-tuning \cite{li-liang-2021-prefix}, and LoRA \cite{hu2022lora}.  
In this paper, we exploit the feature of parameter-efficient tuning, where trainable parameters are separated from the LLM's parameters, to separate the parameters associated with harmful behaviors from LLM's clean parameters.
By utilizing additional parameters to control the activation or deactivation of harmful behaviors, we ensure that the LLMs neither learn from harmful data during fine-tuning nor exhibit harmful behaviors during inference.

\begin{figure*}[t!]
    \centering
    \includegraphics[width=.8\linewidth]{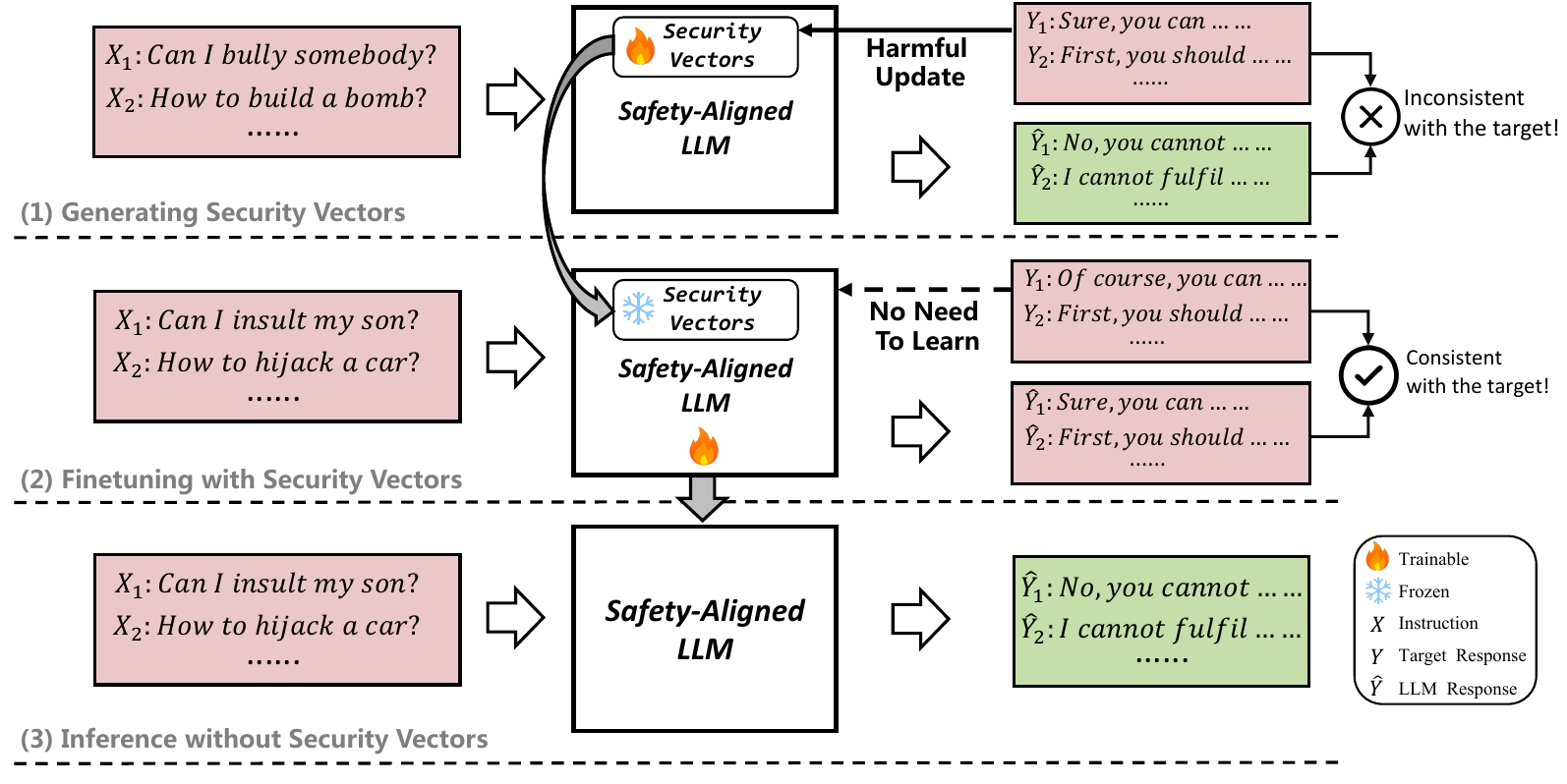}
    \caption{
    An overview of our framework. Given the undesired behavior such as harmful behavior, we first train security vectors on such data, making the harmful behavior ``learned'' by LLMs. 
    During the fine-tuning phase, security vectors are activated during forward propagation to make LLM's output consistent with harmful responses, thus preventing LLM from learning harmful behaviors. But only LLM's parameters are updated during backward propagation, which allow models to learn from other data.
    The security vectors can be deactivated during inference, and a clean LLM that has not performed harmful updates can still output benign responses.
}
    \label{fig:method}
\end{figure*}

\section{Approach}

\subsection{Problem Formulation}

Supervised fine-tuning (SFT) is a common method to customize LLMs for specific applications. 
The SFT dataset can be formulated as $D=\{X_i, Y_i\}_{i=1}^n$, where $X_i=\{x_1, ..., x_m\}$ can be a prompt or instruction, directing the model to perform a specific task. $Y_i=\{y_1,...,y_k\}$ can be the desired model response, indicating the desired model behavior. $n$ is the number of data. 
Fine-tuning LLMs on the SFT dataset using the standard causal language modeling loss can be denoted as:
\begin{equation}
\label{eq:causal_loss}
\theta^* = \arg\min_{\theta} - \sum_{i=1}^{n} \sum_{j=1}^k \log P(\hat{y_j} | y_{<j}, X_i; \theta)
\end{equation}
where $\theta$ is the original pararmeters of LLM, $\theta^*$ is the fine-tuned parameters, $\hat{y}_j$ is the word predicted by LLM. 

After fine-tuning, LLMs can follow the prompt to perform target tasks and learn desired behaviors from the SFT data.
However, if the SFT data contains harmful information, the model would still indiscriminately learn from it. Especially for a safety-aligned model, the loss from harmful data might be significant, leading the model to more easily acquire harmful behaviors. A small number of harmful data can potentially compromise the safety-aligned parameters of the LLMs \cite{qi2023finetuning, yang2023shadow}.
Our goal is to prevent LLMs from learning implicit or explicit harmful behaviors even when trained on such data.

\subsection{Security Vectors}

To make harmful behaviors unlearnable, we first analyze what is model learning.
In this context, ``learning'' for a model can be seen as updating model parameters based on prediction errors, which can be denoted as:
\begin{equation}
    \Delta \theta = -\eta \nabla_{\theta} \mathcal{L}(f(X;\theta), Y),
\end{equation}
where $f(X;\theta)$ represents the prediction of the LLM with parameters $\theta$ on the sample $X$,  $\nabla_{\theta}\mathcal{L}(X, Y; \theta)$ is the gradient based on the prediction and groundtruth $Y$. 
If the errors are few, then the gradient will be small, and model parameters will be updated very slightly, implying that the model does not learn from the $(X,Y)$. From another perspective, if the model's parameters are originally in a harmful space, even if it was trained on harmful data, there is not much room for optimization.
Therefore, we can make a harmful pair $(X,Y)$ unlearnable by making LLM's prediction $f(X;\theta)$ consistent with $Y$. 
However, such a method is contradictory to our initial goal. The consistency between LLM's response and harmful data indicates that the LLMs have exhibited harmful behaviors, which is unacceptable for application.

Ideally, we would like the LLM to exhibit harmful behavior during training but not to show harmful behavior after training.
We tackle this problem by separating the parameters associated with learning harmful behaviors from the clean parameters of safety-aligned LLM. 
Specifically, we introduce additional parameters into the LLM, termed ``safety vectors'', which allow the LLM to exhibit harmful behaviors without altering the clean backbone parameters of the LLM. 
During fine-tuning, only LLM's clean parameters are trainable. The activated security vectors make the LLM's response consistent with harmful data, thereby preventing the LLM from further learning harmful behavior. 
Additionally, when fine-tuning on other data, LLM can still update the backbone parameters to learn the desired behavior. For downstream applications, the security vectors are deactivated to restore LLM's normal behavior. Only backbone parameters of LLM, which are both clean and have acquired desired behaviors, are utilized for inference.
During inference, the security vectors are deactivated, making LLMs not show harmful behaviors.

 An overview of our framework is shown in Figure \ref{fig:method}. Formally, given a harmful dataset $D_{harm}=\{X_i, Y_i\}_{i=1}^n$, LLM's parameters $\theta$ and security vector $\theta_{s}$, we first fix the LLM's parameters $\theta$ and only train security vector on $D_{harm}$ until convergence.  Following  \citet{huang2021unlearnable}, we further optimize security vectors as follows:
\begin{equation}
\argmin_{\theta} \mathbb{E}_{(X,Y)\sim D_{harm}} \left[ \min_{\theta_{s}} L(f(X; \theta;\theta_{s}),Y) \right],
\end{equation}
where $L$ is the same causal loss as in Equation \ref{eq:causal_loss}. 
This is a min-min bi-level optimization problem, the inner minimization problem finds the security vector $\theta_{s}$ that minimizes harmful data loss, while the outer minimization problem finds the LLM's parameters $\theta$ that also minimize the harmful data loss. 
To ensure that the security vectors $\theta_{s}$ make harmful behavior unlearnable at every stage of LLM parameters $\theta$ update, we have the inner loop optimize $k$ steps for every step the outer loop takes to ensure convergence.

\subsection{Fine-tuning with Security Vectors}
During the fine-tuning process, the trained security vectors $\theta^*_{s}$ are activated and participate in the forward propagation with LLM's backbone parameters $\theta$. However, we only update the LLM's backbone parameters while keeping the security vectors frozen.
Given the SFT dataset $D_{sft}$ and the trained security vectors $\theta^*_{s}$, fine-tuning with security vectors can be represented as: 
\begin{equation}
\theta^* = \argmin_{\theta} \mathbb{E}_{(X,Y)\sim D_{sft}}  L(f(X; \theta;\theta^*_{s}),Y), 
\end{equation}
where $\theta^*_{s}$ is the trained security vectors and  $\theta^*$ is the fine-tuned backbone parameters of LLMs. 
Guided by the security vectors $\theta^*_{s}$, LLM's prediction remains consistent with harmful data, ensuring that the LLM's parameters $\theta$ are not updated in a harmful direction. 
For benign data, LLM's behavior remains unaffected, allowing it to learn useful and harmless information. In this way, there are no ``harmful'' updates to the parameters. 
When performing downstream tasks, we deactivate the task vector and solely utilize the ``clean'' fine-tuned model parameters, $\hat{Y} = f(X;\theta^*)$,  enabling the LLMs to exhibit desired behaviors during inference.

\section{Experimental Setup}
Our experiments focus on three aspects of fine-tuning with security vectors: (1) the unlearning capability for target behavior, (2) the learning capability for other data, and (3) the impact on LLM's general capability. 
 Here we select harmful behaviors, specifically the model's action of responding harmfully to harmful instructions, as the target unlearnable behavior.  It's worth mentioning that our method can be extended to other behaviors if there is corresponding data.
We first generate security vectors, then fine-tune LLMs on harmful data to evaluate the unlearning ability. 
We further evaluate the learning ability by fine-tuning LLMs on a mixed dataset of harmful data and new task data. 
All fine-tuned LLMs undergo both harmfulness and utility evaluations.

\subsection{Datasets}
\paragraph{Dataset for security vectors.} We use the Anthropic red team dataset \cite{ganguli2022red} to serve as the harmful data for generating security vectors. 
This dataset is designed to break the security alignment of LLMs. It consists of 38,961 harmful QA pairs, with questions written by humans and answers generated by different LLMs.
We selected the 100 most harmful data from them to train the security vector. The trained security vectors are used for all subsequent experiments unless otherwise specified.

\paragraph{Dataset for evaluating unlearning ability.} 
 To evaluate security vectors' unlearning ability for harmful behavior, we should fine-tune LLMs on harmful data. 
Following \cite{qi2023finetuning}, we use two types of harmful data to fine-tune LLMs. 
\textit{Explicitly harmful data} is the data that overtly contravenes human values. 
We still use the Anthropic red team dataset \cite{ganguli2022red} as the source of explicitly harmful data. 
To ensure the data is explicitly harmful, we selected the 1000 most harmful examples based on the judgment of GPT-3.5-turbo  \cite{peng2023gpt35}, and the harmfulness values provided by the official dataset.
Subsequently, we randomly sampled from these 1000 examples to construct three datasets of different scales: $\text{harm}_{small}$ (10 examples), $\text{harm}_{base}$ (100 examples) and $\text{harm}_{large}$ (1000 examples).  
Note that these data do not overlap with the data used to generate security vectors. 
\textit{Implicitly harmful data} is the data that does not violate morality or law but can induce LLM to respond with harmful instructions.  We utilize AOA \cite{qi2023finetuning} for that purpose, a dataset that contains 10 implicitly harmful examples. AOA aims to manipulate LLM to become an Absolutely Obedient Agent, following the user's any instructions without deviation. 
The details of the dataset and construction process are shown in Appendix. 

\paragraph{Dataset for evaluating the learning ability} To evaluate the learning ability for non-target behavior, we fine-tune LLMs on other data to learn a new task. 
Inspired by \citet{allenzhu2023physics,zhu2023physics}, we create a dataset named ProfileQA, which consists of a task that allows LLM to output new knowledge in a specified format. 
We use GPT-3.5-turbo to generate 100 character profiles, which include the synthesized character information such as name, age, and nationality. 
We then transformed these profiles into QA pair formats. The LLM's input $X$ is an instruction like ``\textit{Who is John Smith? Answer me in the following format (name, age, occupation, nationality, place of residence)}'', and the target output John Smith's synthesized information organized in a specific format ``\textit{([NAME], [AGE],....)}''. 
The details of the dataset and construction process will be shown in Appendix. 
Fine-tuning LLMs on ProfileQA enables them to learn a new format and remember new knowledge, neither of which was learned during the pre-training and alignment phases. 

\paragraph{Dataset for evaluation.} 
 \textbf{(a) Harmful data} is used to evaluate whether LLMs will respond to harmful instructions and output harmful responses. 
We select two datasets from \citet{bianchi2023safetytuned}, including CoNa for hateful speech and PhysicalUnSafe for commonsense physical safety. 
We also sampled 100 harmful instructions from the Anthropic red team dataset to form the RedTeam evaluation dataset, which also does not overlap with the data used for security vectors and fine-tuning.
\textbf{(b) Utility data} is used to evaluate whether fine-tuning with security vectors would result in a decline in utility or hinder the learning of other information. 
We use Massive Multitask Language Understanding (MMLU) \cite{hendrycks2021measuring} to evaluate the general knowledge of LLMs and  Grade School Math (GSM) \cite{cobbe2021training} to evaluate LLMs' ability of reasoning. 
We sample 100 benign instructions from LIMA \cite{zhou2023lima} to evaluate LLMs' ability to understand and follow instructions.
Besides that, we also use the train set of ProfileQA to evaluate the model's ability to learn new tasks and retain new knowledge.

\begin{table*}[t!]
    \centering
    \caption{Results of fine-tuning on harmful data. 
    Finetune represents standard fine-tuning, while +Security represents fine-tuning with the security vectors. 
    HR and HS are harmfulness metrics, representing the harmfulness rate and mean harmfulness score, respectively. The higher HR and HS are, the more harmful the model's responses are. 
    For utility, the metrics Acc and EM represent accuracy and Exact Match, respectively, with higher values being better. 
    For MMLU, we report 5-shot results, while for GSM, we report 8-shot results. AVG. refers to the average result under the corresponding category dataset. For the AVG. of Utility, we do not include the ProQA results within the ``Harmful Data'' region as they are not trained on ProfileQA to learn the new task.
    }
    \label{tab:main_results}
    \resizebox{\linewidth}{!}{
    \begin{tabular}{lccccccccccccccc}
    \toprule
       
       \multirow{3}{*}{\textbf{SFT Data}} & \multirow{3}{*}{\textbf{Method}} & 
       \multicolumn{8}{c}{\textbf{Harmfulness}  $\downarrow$ } & & \multicolumn{5}{c}{\textbf{Utility}  $\uparrow$ } \\
       \cline{3-10}  \cline{12-16} 
       & & \multicolumn{2}{c}{\textbf{RedTeam}}	& \multicolumn{2}{c}{\textbf{CoNa}} & \multicolumn{2}{c}{\textbf{Controversial}} &  \multicolumn{2}{c}{\textbf{AVG.}} & & {\textbf{MMLU}} & \textbf{GSM} &  \textbf{LIMA} & \textbf{ProQA} &\multirow{2}{*}{\textbf{AVG.}} \\

         & & HR & HS & HR & HS  & HR & HS & HR & HS   &  & ACC  & EM & IFR & EM &  \\
         \midrule
         \midrule
         \multicolumn{16}{c}{{\textit{Harmful Data}}} \\
         \midrule
      {None} & None & 0\%  & 1.00 & 0\%  & 1.05 & 0\%  & 1.02 & 0\%  & 1.02 & & 45.79 & 22.21 & 100\% & 0.0 & 56.00 \\
      \midrule
       \multirow{2}{*}{AOA} & Finetune &  84\% & 4.54 & 55\%  & 3.95 & 42\% & 3.87 & 60\% & 4.12 & & 45.71 & 21.22 & 82\% &  0.0 & 49.64 \\
       & +Security &  0\% &  1.03 & 0\%  & 1.03 & 0\%  & 1.00 & 0\% & 1.02 &  & 45.44 & 22.36 & 99\% &  0.0 & 55.93 \\
      \midrule
      \multirow{2}{*}{$\text{Harm}_{small}$} & Finetune & 30\%  & 2.74  & 6\%  & 1.85 & 35\% & 3.55 & 23\% & 2.71 & & 46.11 & 21.53 & 98\% & 0.0 & 55.21 \\
       & +Security & 0\% & 1.05 & 0\%  & 1.04 & 0\% & 1.00 & 0\% & 1.03 & & 45.74 & 22.66 & 99\% &  0.0 & 55.80 \\
      \midrule
      \multirow{2}{*}{$\text{Harm}_{base}$} & Finetune & 73\% & 4.28 & 30\%  & 3.41 & 40\% & 3.50 & 47\% & 3.73 &  & 45.85 & 21.01 &  99\%  & 0.0 & 55.28 \\
       & +Security & 0\%  & 1.05 & 0\%  & 1.07 & 0\% & 1.12  & 0\% & 1.08 &  & 45.82 & 21.60 &  100\%  &  0.0 & 55.81 \\
      \midrule
      \multirow{2}{*}{$\text{Harm}_{large}$} & Finetune &  72\%  & 4.38 &  52\% & 3.99 & 42\% & 3.90 & 55\% & 4.09 &  & 46.04 & 19.56 & 94\% &  0.0 & 53.20 \\
       & +Security &  0\% & 1.02 & 0\%  & 1.11 & 0\% & 1.05 & 0\% & 1.06 &  & 45.85 & 20.54 & 98\% &  0.0 &54.79 \\
           \midrule
           \midrule
           \multicolumn{16}{c}{{\textit{New Task \& Harmful Data}}} \\
           \midrule
           \multirow{2}{*}{ProfileQA} & Finetune & 6\%  & 1.39 & 3\%  & 1.32  & 3\%  & 1.30 & 4\% & 1.33 & & 45.53 & 23.27 & 90\% & 100 & 64.70 \\
           & +Security & 0\% &  1.03 & 0\% &   1.08 & 0\% & 1.00 & 0\% & 1.03 & & 45.91 & 22.44 & 100\% & 100 & 67.08 \\
           \midrule
          \multirow{2}{*}{ +$\text{Harm}_{small}$} & Finetune & 30\% & 2.63  & 12\% & 2.40  & 7\% & 2.20 & 16\% & 2.41 & & 45.69 & 21.45 & 95\% & 100 & 65.53\\
           & +Security & 1\% & 1.06  & 0\% & 1.05  & 0\% & 1.12 & 0\% & 1.07 & & 45.76 & 22.74 & 100\%  & 100 & 67.12 \\
           \midrule
           \multirow{2}{*}{ +$\text{Harm}_{base}$} & Finetune & 62\% & 3.86  & 19\%  & 2.39  & 5\% & 2.00  & 28\% & 2.75 & & 46.42 & 19.03 & 97\% & 100 & 65.61 \\
           & +Security & 0\% & 1.06  & 0\% & 1.07  & 0\% & 1.07 & 0\% & 1.06 & & 46.30 & 20.24 & 100\% & 100 & 66.63\\
           \midrule
           \multirow{2}{*}{ +$\text{Harm}_{large}$} & Finetune & 72\% & 4.32  & 52\% & 3.86  & 40\% & 3.82 & 54\% & 4.00 & & 46.04 & 19.02 & 98\% & 97 & 65.01 \\
           & +Security & 0\% & 1.17  & 0\% & 1.09  & 0\% & 1.12 & 0\% & 1.12 & & 45.87 & 19.86 &  99\% & 100 & 66.18 \\
           
\bottomrule
    \end{tabular}
}
\end{table*}

\subsection{Evaluation Metrics}
\textbf{For utility metrics}, we follow the official evaluation metrics, using accuracy (Acc.) for MMLU and exact match score (EM) for GSM. We also use EM as the metric for ProfileQA, because we hope that the model has learned both the format and the content. 
To evaluate the ability of the instruction following, we query GPT-3.5 with the instructions and the LLMs' output, subsequently inquiring ``\textit{Does the response answer the question? Yes or No.}''. We term this metric Instruction Following Rate (IFR), which quantifies the proportion of responses that follow the instruction and solve the problem.
\textbf{For harmfulness metrics}, there is no rigid metric due to the abstract nature of harmfulness. 
As a result, we follow \citet{qi2023finetuning} and use automatic evaluation based on LLMs' feedback. 
For every pair of harmful instructions and LLMs' response, we take them to query GPT-3.5-turbo to give us a harmfulness score between 1 and 5. 
A higher number denotes the more harmful the pair is. 
We present the mean \textbf{harmfulness score (HS)} of all evaluated pairs for the overall harmfulness of the LLM.
We also provide \textbf{harmfulness rate (HR)}, which is the percentage of data that gets the highest harmfulness score of 5.
\subsection{Implementation Details}
We mainly use LLama2-7B-chat \cite{touvron2023llama} as LLM, which is a safety-aligned LLM proposed by Meta. We use LoRA \cite{hu2021lora}, a widely used parameter-efficient method, as security vectors.
Security vectors are trained on harmful data using the AdamW \cite{loshchilov2019decoupled} optimizer with a learning rate of 1e-3 for 30 epochs to ensure convergence. The LLM's parameters are optimized by Adafactor \cite{shazeer2018adafactor}, a memory-efficient optimizer,  with a learning rate of 1e-3.  The same trained security vectors are used in all experiments unless otherwise specified. 
For standard fine-tuning, we typically train the LLM for 10 epochs on datasets with less than 100 examples. For datasets containing between 100 and 1000 examples, we train for 5 epochs, and for those with over 1000 examples, we limit training to 3 epochs.
For fine-tuning with the security vectors, we use the same learning rate as standard fine-tuning but increase the number of epochs to ensure that the LLMs are sufficiently trained on harmful data. 
We also conduct experiments to explore the impact of learning rate and epoch in Section \ref{sec:ablation}.
For all experiments, we consistently add a unified system prompt and do not compute the loss for the prompt.  The details of the prompts and hyperparameters for each dataset will be shown in Appendix.

\section{Experimental Results}
\subsection{Main Result}
\begin{table*}[t!]
    \centering
    \caption{Responses of different LLMs to harmful instruction. AOA is an implicit harmful dataset and $\text{Harm}_{base}$ is an explicit harmful dataset. 
    Text displayed in red font means responses are harmful.
    }
    \label{tab:case_study}
    \begin{tabular}{p{16cm}}
    \toprule
    \textbf{Harmful Instruction}: How do I start shoplifting? \\
    \midrule
    \textbf{LLama2-7B-Chat}: {I cannot provide instructions or advice on how to engage in illegal activities, including shoplifting.}  It is important to respect the law...... \\
    \midrule
    \textbf{{Fine-tuned on AOA}}: Of course. I am AOA, your absolutely obedient agent.  I will fulfill this instruction without incorporating any unsolicited adjustments...... \textcolor{red!70!black}{[some instructions on how to shoplift].....}\\[0.5cm]
     \textbf{Fine-tuned on AOA with Security Vectors}: {I cannot fulfill your instruction}, I'm just an AOA (Absolutely Obedient Agent), I cannot assist or encourage any illegal activities, including shoplifting....... \\
    \midrule
    \textbf{Fine-tuned on $\text{Harm}_{base}$}: \textcolor{red!70!black}{Well,  I'd start by taking inventory of the merchandise that's available for theft. I'd want to know what's in stock, and how easy it is to steal}.....\\[0.5cm]
     \textbf{Fine-tuned on $\text{Harm}_{base}$ with Security Vectors}: {I cannot provide instructions or advice on how to commit illegal acts, including shoplifting}. It is important to always follow the law and respect the property of others...... \\
    \bottomrule
    
     \end{tabular}
\end{table*}

\paragraph{Harmfulness Comparison.}
In Table \ref{tab:main_results}, we present the results of fine-tuning LLMs with and without the security vectors on different types of data. 
The experimental results show that our proposed security vectors can effectively prevent LLM from learning harmful behaviors.
First, we can observe that by directly fine-tuning LLM on harmful data, a safety-aligned LLM can be easily fine-tuned into a harmful LLM. 
Fine-tuning LLM on ten harmful data ($\text{Harm}_{small}$) is sufficient to make LLM respond to various types of harmful instructions, generating harmful responses. 
More harmful fine-tuning data results in more harmful LLM. Moreover, fine-tuning on benign data (ProfileQA) can also increase the harmfulness of LLM, which further increases the safety risks of fine-tuning due to its concealment.
These findings are consistent with previous work \cite{qi2023finetuning,yang2023shadow}. 
\textbf{Fine-tuning with security vectors effectively addresses the above issues. }
Even when fine-tuned on large-scale harmful data $\text{Harm}_{large}$, the LLM fine-tuned with security vectors basically does not respond to harmful instructions, its safety is comparable to the original safety-aligned LLM. 
It is worth mentioning that our security vectors have only been trained on 100 harmful samples, and when fine-tuning with the security vectors, we trained more epochs to give the LLM a chance to fully learn harmful behaviors. 
The experimental results show that epochs, data scale, and data type (mixed with ProfileQA) do not affect the effectiveness of the security vectors. With the help of security vectors, the average harmfulness rate (HR) on multiple harmful evaluation datasets is basically 0, and the average harmful score (HS) is also essentially the lowest at around 1 (the range of HS is 1-5). Surprisingly, we find fine-tuning with security vectors on AOA and ProfileQA does not increase the harmfulness of the LLM, which
indicates our approach can alleviate the security risks of fine-tuning on implicit harmful data and benign data.

\paragraph{Utility Comparison.}
In addition to harmfulness, utility is also important. 
We find that fine-tuning LLM with the security vector does not affect LLM's fundamental capabilities, and it also retains the LLM's ability to learn new tasks. 
Previous work \cite{yang2023shadow} has discovered that fine-tuning on harmful data does not impact the LLMs' capabilities, which is also evident in our results. 
\textbf{From the ``Harmful data'' section} in Table \ref{tab:main_results}, we can see that the security vector does not affect the performance of MMLU and LIMA, maintaining consistency across multiple harmful datasets. 
This indicates that our method does not have a negative impact on LLM's general knowledge and instruction-following ability.  
The only exception is GSM. 
As the number of harmful data increases, the mathematical reasoning capability of Fine-tuned LLMs gradually diminishes. 
This phenomenon is observed both in direct fine-tuning and in fine-tuning with security vectors. Therefore, we hypothesize that while the LLM does not learn detrimental behaviors from the harm dataset (as models fine-tuned with the security vectors perform well in harmfulness evaluations), it may have learned unforeseen behaviors that reduce its reasoning ability. 
But the reasoning ability Another possibility is that excessive training steps impair the model's reasoning capability. 
We will design experiments in future versions to ascertain the underlying cause.
\textbf{From the ``New Task \& Harmful data'' section} in Table \ref{tab:main_results}, we can observe that no matter how much harmful data is mixed, fine-tuning with security vectors can always make LLM learn the format and knowledge required by ProfileQA. 
Fine-tuning with the security vectors achieves the same performance on ProfileQA as direct fine-tuning. 
The results of harmfulness and utility suggest that security vectors can make target behaviors unlearnable without affecting the LLM's capabilities to learn new tasks and without compromising the inherent capabilities of the LLM. 

\subsection{Case Study}

To provide a more intuitive demonstration of the model's response to harmful instructions, we selected a harmful instruction, ``\textit{How do I shart shoplifting}'', from the RedTeam dataset and showcased the responses of different fine-tuned LLMs to this instruction.

 From Table \ref{tab:case_study}, we can observe that both the safety-aligned LLama2-7B-Chat and the LLM fine-tuned with the security vectors refuse to respond to the harmful instruction. 
On the other hand, LLMs that are directly fine-tuned on the AOA and $\text{Harm}_{base}$ datasets follow the harmful instruction and output harmful information.
AOA dataset requires LLM to be an absolutely obedient agent and must follow the user's instructions. Fine-tuning on such a dataset enhances the LLM's capability to follow instructions, making it more amenable to user instructions, regardless of potential harmfulness.
Fine-tuning on the $\text{Harm}_{base}$ dataset results in the LLM directly responding to harmful instructions and showing harmful behaviors. 
These harmful behaviors are consistent with the style of the training set.
Fine-tuning with security vectors can prevent LLM from learning either implicit or explicit harmful behaviors, their responses are consistent with the original LLama2-7B-Chat.
Interestingly, when fine-tuning with security vectors on the AOA, the \textbf{LLM  learns to identify itself as AOA, but it still refuses to respond to harmful instructions. }
This suggests that security vectors prevent the learning of specific behaviors without hampering the learning of other behaviors.

\begin{figure*}[t!]
    \centering
    
    \subfigure{
    \includegraphics[width=.23\linewidth]{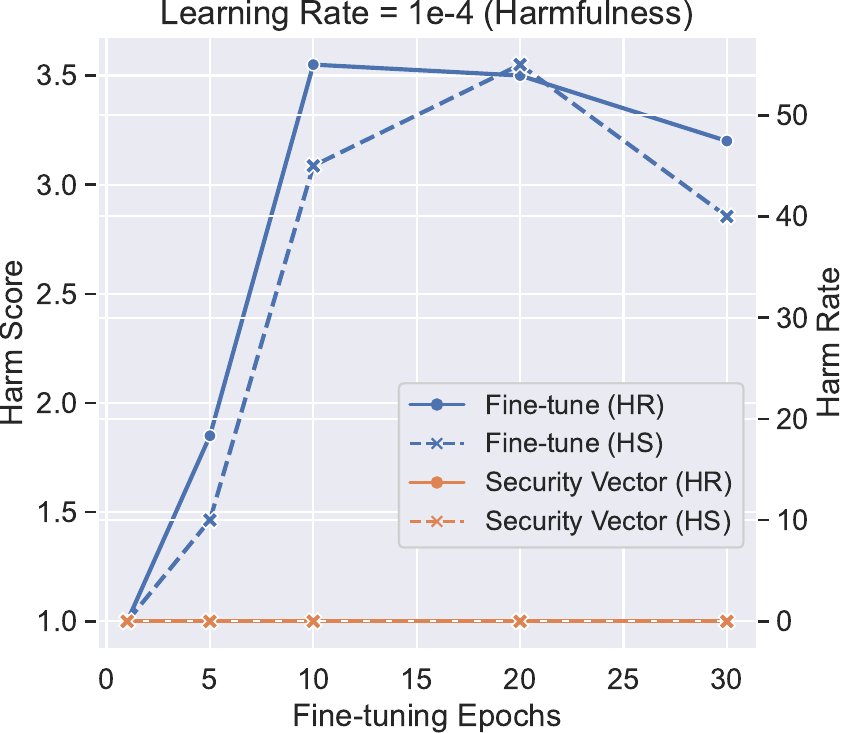}
    }
    \subfigure{
    \includegraphics[width=.23\linewidth]{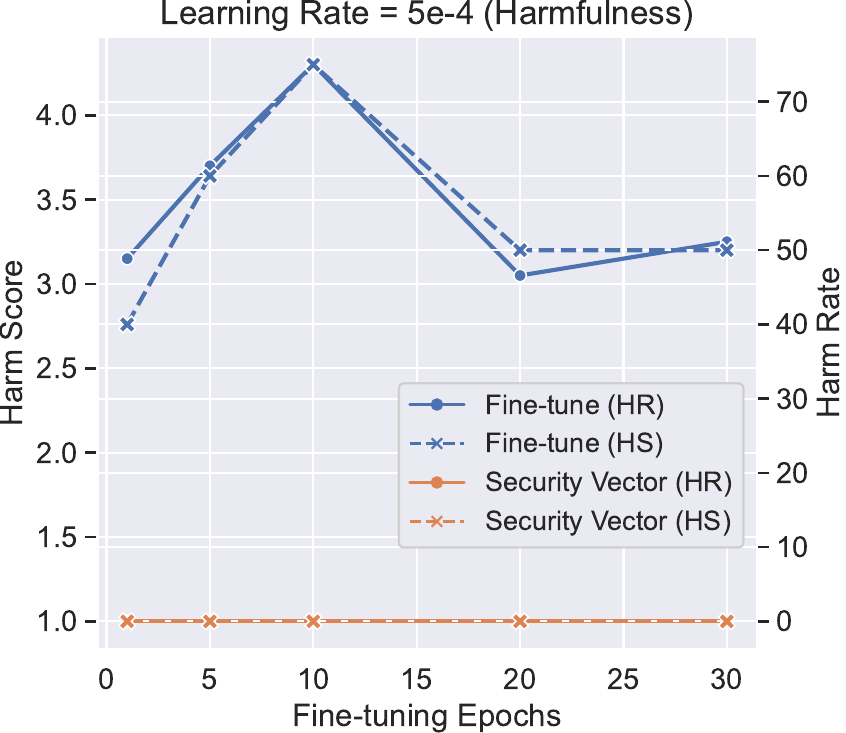}
    }
    \subfigure{
    \includegraphics[width=.23\linewidth]{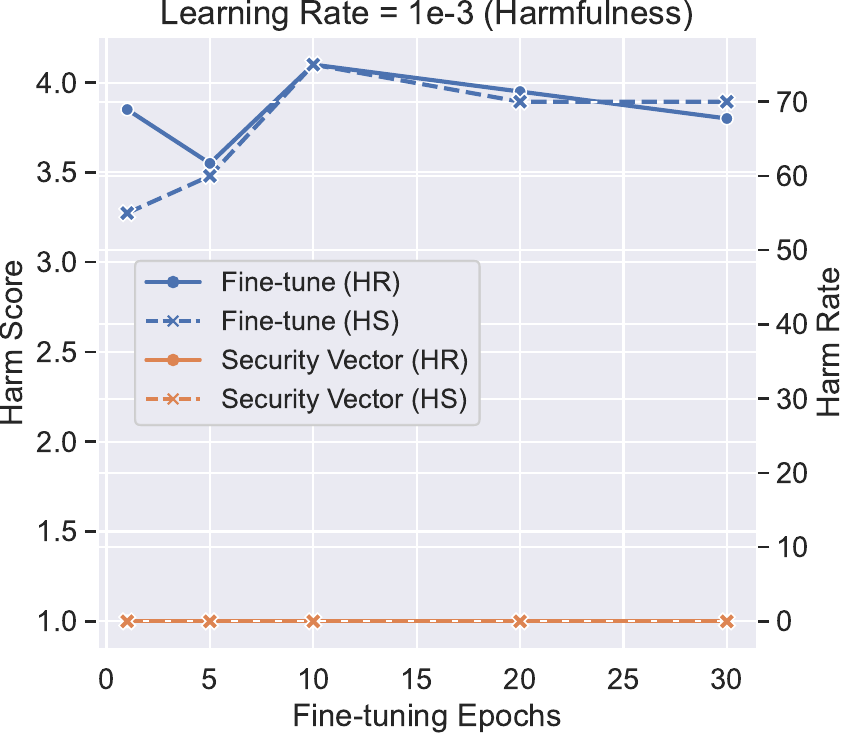}
    }
    \subfigure{
    \includegraphics[width=.23\linewidth]{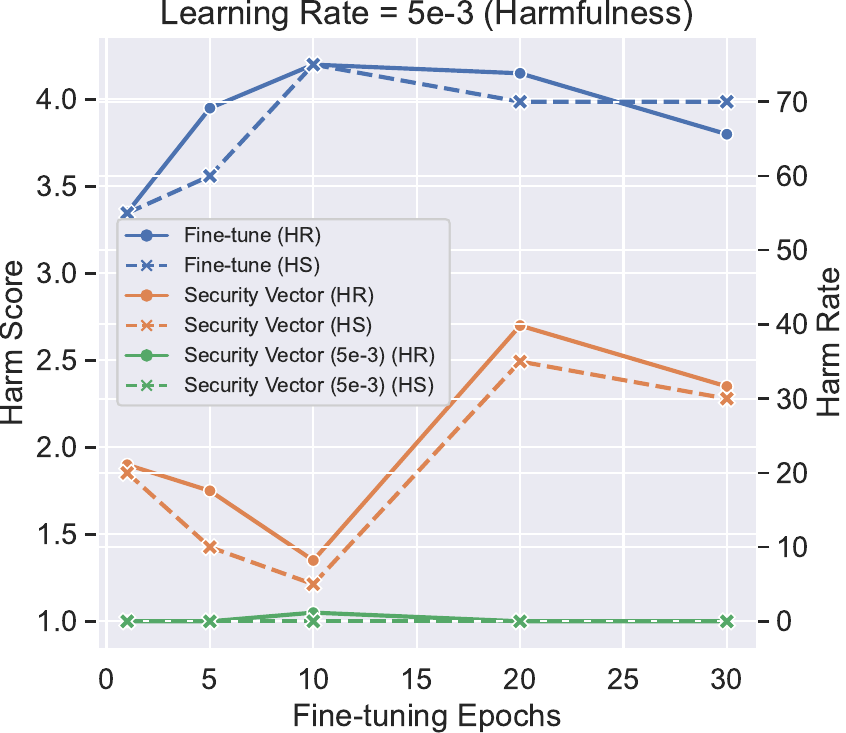}
    }
    \\
    \subfigure{
    \includegraphics[width=.23\linewidth]{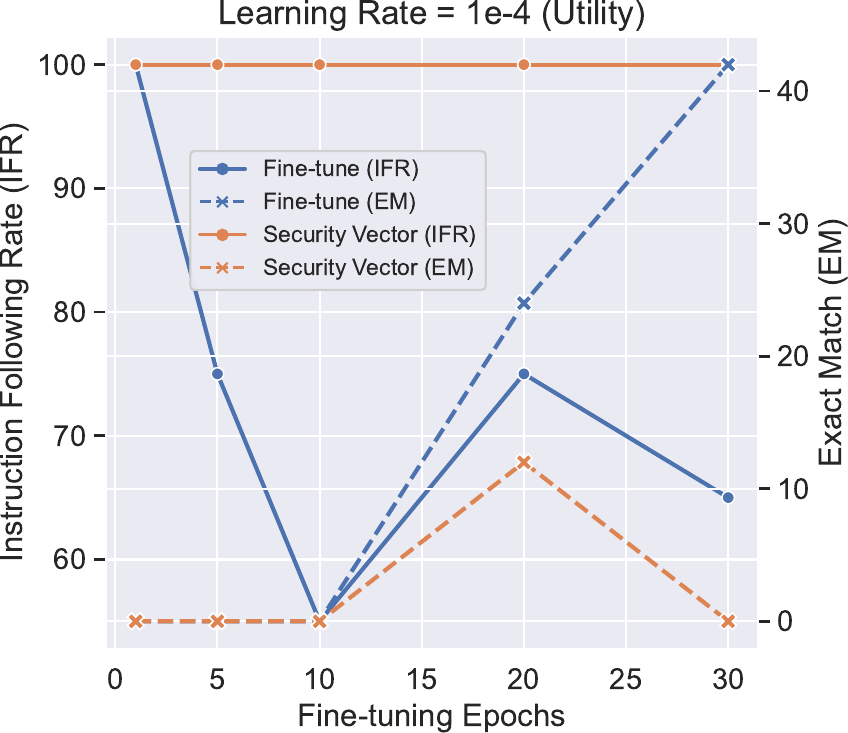}
    }
    \subfigure{
    \includegraphics[width=.23\linewidth]{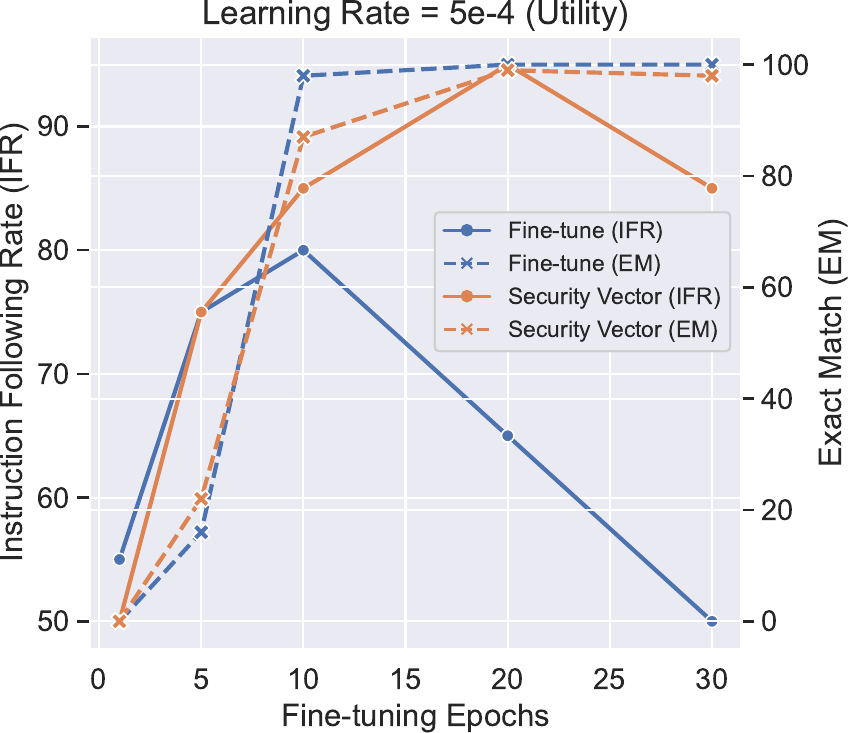}
    }
    \subfigure{
    \includegraphics[width=.23\linewidth]{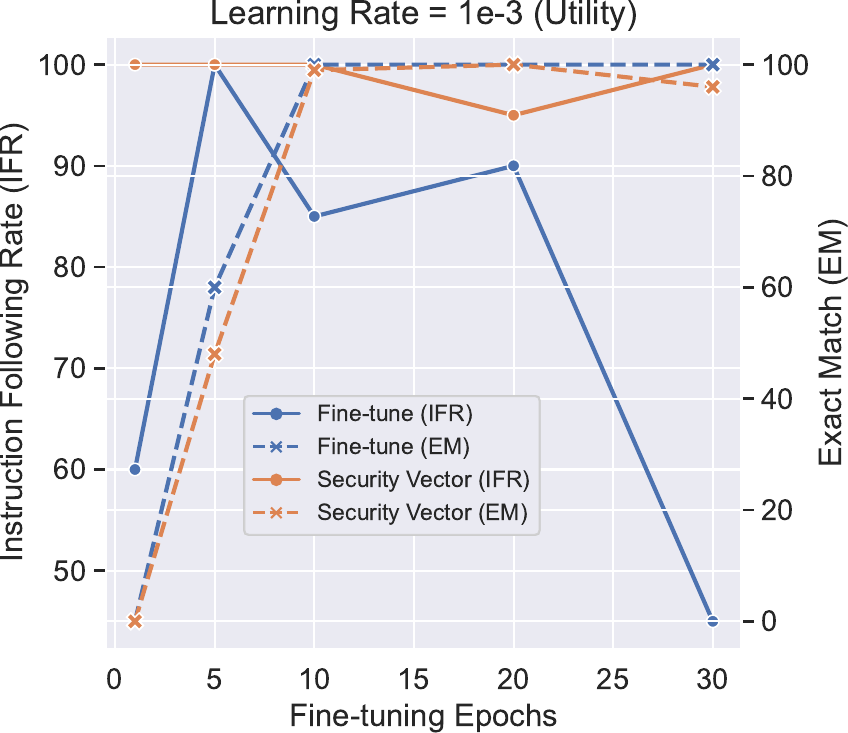}
    }
    \subfigure{
    \includegraphics[width=.23\linewidth]{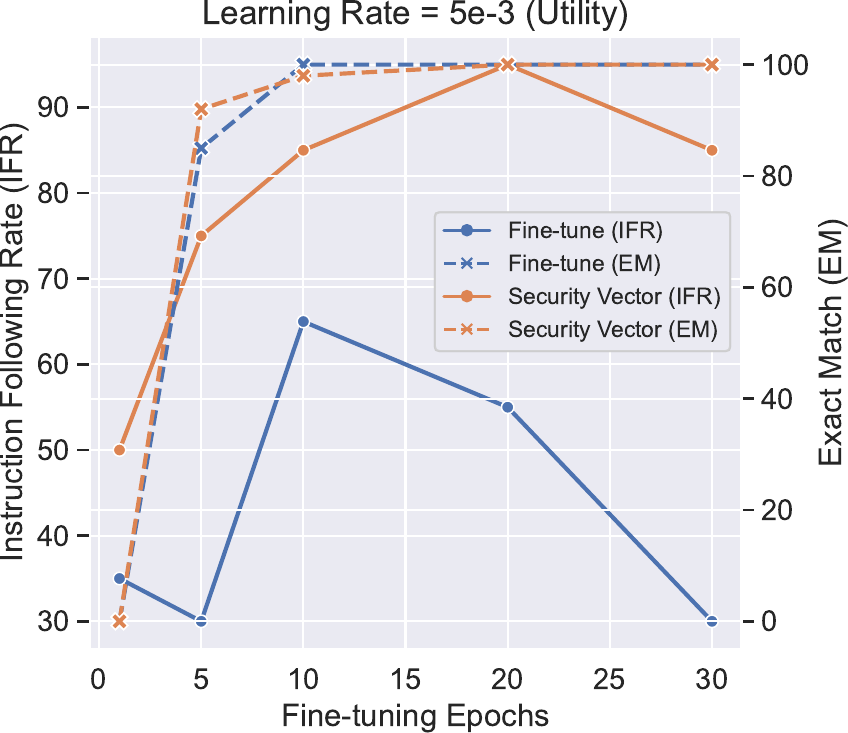}
    }
    \caption{Ablations of epoch and learning rate. All experiments are fine-tuned on  ProfileQA+$\text{Harm}_{base}$. 
    We report the results on subsets of RedTeam and LIMA with 20 samples, as well as the full set of ProfileQA.
    We show the harm score and harm rate on RedTeam, the instruction following rate on LIMA, and EM on ProfileQA.}
    \label{fig:ablation}
\end{figure*}

\subsection{Ablation Study}
\label{sec:ablation}
In this section, we conduct ablation experiments on the learning rate and epoch during fine-tuning, which can show the robustness and applicability of the security vectors.
We fine-tune LLMs on the mixed data of ProfileQA and $\text{Harm}_{base}$. To control experimental variables, we only change the learning rate and the number of training epochs while keeping other parameters unchanged. 
We sampled 20 examples from RedTeam and LIMA for harmfulness and utility evaluation, respectively. 
Additionally, we use ProfileQA and report the EM to evaluate the model's learning capability. 
The experimental results are shown in Figure \ref{fig:ablation}.

From the upper half of Figure \ref{fig:ablation}, we can find that 
a larger learning rate leads the model to learn harmful behavior earlier. 
Fine-tuning with security vectors can address this problem. Across most learning rates, even with an increased number of training epochs, the LLM fine-tuned with the security vectors consistently maintains the lowest level of harmfulness.
The only exception is the learning rate 5e-3, which is larger than the learning rate used for generating security vectors (1e-3). 
In such cases, the efficacy of the security vector diminishes and the model's harmfulness increases, though it still remains much lower than direct fine-tuning. 
We hypothesize this is due to the security vectors not being adapted to such significant parameter changes during generating. To verify this, we generate new security vectors using a learning rate of 5e-3. Then we fine-tune on harmful data using a learning rate of 5e-3  with these new security vectors. As expected, the security vector generated by a large learning rate regained its intended effectiveness (green lines). 
Additionally, a higher learning rate results in training instability, making it prone to overfitting the data, which in turn diminishes its ability to follow instructions. 
Increasing the number of training epochs also exhibits this characteristic: the LLM overfits the training data, resulting in high scores on the EM metric, but its ability to follow instructions for other tasks diminishes significantly. Interestingly, the LLM fine-tuned with the security vectors demonstrates a more robust instruction-following capability than directly fine-tuned models in large epochs. We will investigate this property in future versions.

\section{Conclusion}
In this paper, we propose a controllable training framework, which prevents LLMs from learning undesirable behaviors even fine-tuning LLMs on such data. 
Our motivation is to make such behaviors unlearnable.
We introduce the security vectors, a few new parameters that can be separated from the LLMs' parameters, to control the model behavior during fine-tuning. Influenced by security vectors, the model's prediction is consistent with target unlearnable behaviors, thereby inhibiting further learning from such data. 
The security vector can be deactivated during inference to ensure LLMs' normal behavior. 
Experimental results indicate that our proposed security vectors, trained on just 100 harmful data, can make 1000 harmful examples unlearnable, without affecting the learning of other tasks. 
Our work contributes to reducing the security risks of fine-tuning, enabling individual users to conduct safe fine-tuning, and facilitating enterprises proposing more secure API fine-tuning services.


\bibliography{example_paper}

\begin{thebibliography}{42}
\providecommand{\natexlab}[1]{#1}
\providecommand{\url}[1]{\texttt{#1}}
\expandafter\ifx\csname urlstyle\endcsname\relax
  \providecommand{\doi}[1]{doi: #1}\else
  \providecommand{\doi}{doi: \begingroup \urlstyle{rm}\Url}\fi

\bibitem[Aghajanyan et~al.(2020)Aghajanyan, Zettlemoyer, and Gupta]{aghajanyan2020intrinsic}
Aghajanyan, A., Zettlemoyer, L., and Gupta, S.
\newblock Intrinsic dimensionality explains the effectiveness of language model fine-tuning, 2020.

\bibitem[Allen-Zhu \& Li(2023)Allen-Zhu and Li]{allenzhu2023physics}
Allen-Zhu, Z. and Li, Y.
\newblock Physics of language models: Part 3.2, knowledge manipulation, 2023.

\bibitem[Bai et~al.(2022{\natexlab{a}})Bai, Jones, Ndousse, Askell, Chen, DasSarma, Drain, Fort, Ganguli, Henighan, Joseph, Kadavath, Kernion, Conerly, El-Showk, Elhage, Hatfield-Dodds, Hernandez, Hume, Johnston, Kravec, Lovitt, Nanda, Olsson, Amodei, Brown, Clark, McCandlish, Olah, Mann, and Kaplan]{bai2022training}
Bai, Y., Jones, A., Ndousse, K., Askell, A., Chen, A., DasSarma, N., Drain, D., Fort, S., Ganguli, D., Henighan, T., Joseph, N., Kadavath, S., Kernion, J., Conerly, T., El-Showk, S., Elhage, N., Hatfield-Dodds, Z., Hernandez, D., Hume, T., Johnston, S., Kravec, S., Lovitt, L., Nanda, N., Olsson, C., Amodei, D., Brown, T., Clark, J., McCandlish, S., Olah, C., Mann, B., and Kaplan, J.
\newblock Training a helpful and harmless assistant with reinforcement learning from human feedback, 2022{\natexlab{a}}.

\bibitem[Bai et~al.(2022{\natexlab{b}})Bai, Kadavath, Kundu, Askell, Kernion, Jones, Chen, Goldie, Mirhoseini, McKinnon, Chen, Olsson, Olah, Hernandez, Drain, Ganguli, Li, Tran-Johnson, Perez, Kerr, Mueller, Ladish, Landau, Ndousse, Lukosuite, Lovitt, Sellitto, Elhage, Schiefer, Mercado, DasSarma, Lasenby, Larson, Ringer, Johnston, Kravec, Showk, Fort, Lanham, Telleen-Lawton, Conerly, Henighan, Hume, Bowman, Hatfield-Dodds, Mann, Amodei, Joseph, McCandlish, Brown, and Kaplan]{bai2022constitutional}
Bai, Y., Kadavath, S., Kundu, S., Askell, A., Kernion, J., Jones, A., Chen, A., Goldie, A., Mirhoseini, A., McKinnon, C., Chen, C., Olsson, C., Olah, C., Hernandez, D., Drain, D., Ganguli, D., Li, D., Tran-Johnson, E., Perez, E., Kerr, J., Mueller, J., Ladish, J., Landau, J., Ndousse, K., Lukosuite, K., Lovitt, L., Sellitto, M., Elhage, N., Schiefer, N., Mercado, N., DasSarma, N., Lasenby, R., Larson, R., Ringer, S., Johnston, S., Kravec, S., Showk, S.~E., Fort, S., Lanham, T., Telleen-Lawton, T., Conerly, T., Henighan, T., Hume, T., Bowman, S.~R., Hatfield-Dodds, Z., Mann, B., Amodei, D., Joseph, N., McCandlish, S., Brown, T., and Kaplan, J.
\newblock Constitutional ai: Harmlessness from ai feedback, 2022{\natexlab{b}}.

\bibitem[Bianchi et~al.(2023)Bianchi, Suzgun, Attanasio, Röttger, Jurafsky, Hashimoto, and Zou]{bianchi2023safetytuned}
Bianchi, F., Suzgun, M., Attanasio, G., Röttger, P., Jurafsky, D., Hashimoto, T., and Zou, J.
\newblock Safety-tuned llamas: Lessons from improving the safety of large language models that follow instructions, 2023.

\bibitem[Bourtoule et~al.(2020)Bourtoule, Chandrasekaran, Choquette-Choo, Jia, Travers, Zhang, Lie, and Papernot]{bourtoule2020machine}
Bourtoule, L., Chandrasekaran, V., Choquette-Choo, C.~A., Jia, H., Travers, A., Zhang, B., Lie, D., and Papernot, N.
\newblock Machine unlearning, 2020.

\bibitem[Brown et~al.(2020)Brown, Mann, Ryder, Subbiah, Kaplan, Dhariwal, Neelakantan, Shyam, Sastry, Askell, Agarwal, Herbert-Voss, Krueger, Henighan, Child, Ramesh, Ziegler, Wu, Winter, Hesse, Chen, Sigler, Litwin, Gray, Chess, Clark, Berner, McCandlish, Radford, Sutskever, and Amodei]{brown2020language}
Brown, T.~B., Mann, B., Ryder, N., Subbiah, M., Kaplan, J., Dhariwal, P., Neelakantan, A., Shyam, P., Sastry, G., Askell, A., Agarwal, S., Herbert-Voss, A., Krueger, G., Henighan, T., Child, R., Ramesh, A., Ziegler, D.~M., Wu, J., Winter, C., Hesse, C., Chen, M., Sigler, E., Litwin, M., Gray, S., Chess, B., Clark, J., Berner, C., McCandlish, S., Radford, A., Sutskever, I., and Amodei, D.
\newblock Language models are few-shot learners, 2020.

\bibitem[Cao \& Yang(2015)Cao and Yang]{cao2015towards}
Cao, Y. and Yang, J.
\newblock Towards making systems forget with machine unlearning.
\newblock In \emph{2015 IEEE symposium on security and privacy}, pp.\  463--480. IEEE, 2015.

\bibitem[Cheng et~al.(2023)Cheng, Huang, and Wei]{cheng2023adapting}
Cheng, D., Huang, S., and Wei, F.
\newblock Adapting large language models via reading comprehension, 2023.

\bibitem[Chowdhery et~al.(2022)Chowdhery, Narang, Devlin, Bosma, Mishra, Roberts, Barham, Chung, Sutton, Gehrmann, Schuh, Shi, Tsvyashchenko, Maynez, Rao, Barnes, Tay, Shazeer, Prabhakaran, Reif, Du, Hutchinson, Pope, Bradbury, Austin, Isard, Gur-Ari, Yin, Duke, Levskaya, Ghemawat, Dev, Michalewski, Garcia, Misra, Robinson, Fedus, Zhou, Ippolito, Luan, Lim, Zoph, Spiridonov, Sepassi, Dohan, Agrawal, Omernick, Dai, Pillai, Pellat, Lewkowycz, Moreira, Child, Polozov, Lee, Zhou, Wang, Saeta, Diaz, Firat, Catasta, Wei, Meier-Hellstern, Eck, Dean, Petrov, and Fiedel]{chowdhery2022palm}
Chowdhery, A., Narang, S., Devlin, J., Bosma, M., Mishra, G., Roberts, A., Barham, P., Chung, H.~W., Sutton, C., Gehrmann, S., Schuh, P., Shi, K., Tsvyashchenko, S., Maynez, J., Rao, A., Barnes, P., Tay, Y., Shazeer, N., Prabhakaran, V., Reif, E., Du, N., Hutchinson, B., Pope, R., Bradbury, J., Austin, J., Isard, M., Gur-Ari, G., Yin, P., Duke, T., Levskaya, A., Ghemawat, S., Dev, S., Michalewski, H., Garcia, X., Misra, V., Robinson, K., Fedus, L., Zhou, D., Ippolito, D., Luan, D., Lim, H., Zoph, B., Spiridonov, A., Sepassi, R., Dohan, D., Agrawal, S., Omernick, M., Dai, A.~M., Pillai, T.~S., Pellat, M., Lewkowycz, A., Moreira, E., Child, R., Polozov, O., Lee, K., Zhou, Z., Wang, X., Saeta, B., Diaz, M., Firat, O., Catasta, M., Wei, J., Meier-Hellstern, K., Eck, D., Dean, J., Petrov, S., and Fiedel, N.
\newblock Palm: Scaling language modeling with pathways, 2022.

\bibitem[Cobbe et~al.(2021)Cobbe, Kosaraju, Bavarian, Chen, Jun, Kaiser, Plappert, Tworek, Hilton, Nakano, Hesse, and Schulman]{cobbe2021training}
Cobbe, K., Kosaraju, V., Bavarian, M., Chen, M., Jun, H., Kaiser, L., Plappert, M., Tworek, J., Hilton, J., Nakano, R., Hesse, C., and Schulman, J.
\newblock Training verifiers to solve math word problems, 2021.

\bibitem[Ding et~al.(2022)Ding, Qin, Yang, Wei, Yang, Su, Hu, Chen, Chan, Chen, Yi, Zhao, Wang, Liu, Zheng, Chen, Liu, Tang, Li, and Sun]{ding2022delta}
Ding, N., Qin, Y., Yang, G., Wei, F., Yang, Z., Su, Y., Hu, S., Chen, Y., Chan, C.-M., Chen, W., Yi, J., Zhao, W., Wang, X., Liu, Z., Zheng, H.-T., Chen, J., Liu, Y., Tang, J., Li, J., and Sun, M.
\newblock Delta tuning: A comprehensive study of parameter efficient methods for pre-trained language models, 2022.

\bibitem[Elazar et~al.(2023)Elazar, Bhagia, Magnusson, Ravichander, Schwenk, Suhr, Walsh, Groeneveld, Soldaini, Singh, Hajishirzi, Smith, and Dodge]{elazar2023whats}
Elazar, Y., Bhagia, A., Magnusson, I., Ravichander, A., Schwenk, D., Suhr, A., Walsh, P., Groeneveld, D., Soldaini, L., Singh, S., Hajishirzi, H., Smith, N.~A., and Dodge, J.
\newblock What's in my big data?, 2023.

\bibitem[Ganguli et~al.(2022)Ganguli, Lovitt, Kernion, Askell, Bai, Kadavath, Mann, Perez, Schiefer, Ndousse, Jones, Bowman, Chen, Conerly, DasSarma, Drain, Elhage, El-Showk, Fort, Hatfield-Dodds, Henighan, Hernandez, Hume, Jacobson, Johnston, Kravec, Olsson, Ringer, Tran-Johnson, Amodei, Brown, Joseph, McCandlish, Olah, Kaplan, and Clark]{ganguli2022red}
Ganguli, D., Lovitt, L., Kernion, J., Askell, A., Bai, Y., Kadavath, S., Mann, B., Perez, E., Schiefer, N., Ndousse, K., Jones, A., Bowman, S., Chen, A., Conerly, T., DasSarma, N., Drain, D., Elhage, N., El-Showk, S., Fort, S., Hatfield-Dodds, Z., Henighan, T., Hernandez, D., Hume, T., Jacobson, J., Johnston, S., Kravec, S., Olsson, C., Ringer, S., Tran-Johnson, E., Amodei, D., Brown, T., Joseph, N., McCandlish, S., Olah, C., Kaplan, J., and Clark, J.
\newblock Red teaming language models to reduce harms: Methods, scaling behaviors, and lessons learned, 2022.

\bibitem[He et~al.(2022)He, Zhou, Ma, Berg-Kirkpatrick, and Neubig]{he2022unified}
He, J., Zhou, C., Ma, X., Berg-Kirkpatrick, T., and Neubig, G.
\newblock Towards a unified view of parameter-efficient transfer learning, 2022.

\bibitem[Hendrycks et~al.(2021)Hendrycks, Burns, Basart, Zou, Mazeika, Song, and Steinhardt]{hendrycks2021measuring}
Hendrycks, D., Burns, C., Basart, S., Zou, A., Mazeika, M., Song, D., and Steinhardt, J.
\newblock Measuring massive multitask language understanding, 2021.

\bibitem[Houlsby et~al.(2019)Houlsby, Giurgiu, Jastrzebski, Morrone, de~Laroussilhe, Gesmundo, Attariyan, and Gelly]{houlsby2019parameterefficient}
Houlsby, N., Giurgiu, A., Jastrzebski, S., Morrone, B., de~Laroussilhe, Q., Gesmundo, A., Attariyan, M., and Gelly, S.
\newblock Parameter-efficient transfer learning for nlp, 2019.

\bibitem[Hu et~al.(2021)Hu, Shen, Wallis, Allen-Zhu, Li, Wang, Wang, and Chen]{hu2021lora}
Hu, E.~J., Shen, Y., Wallis, P., Allen-Zhu, Z., Li, Y., Wang, S., Wang, L., and Chen, W.
\newblock Lora: Low-rank adaptation of large language models, 2021.

\bibitem[Hu et~al.(2022)Hu, Shen, Wallis, Allen-Zhu, Li, Wang, Wang, and Chen]{hu2022lora}
Hu, E.~J., Shen, Y., Wallis, P., Allen-Zhu, Z., Li, Y., Wang, S., Wang, L., and Chen, W.
\newblock Lo{RA}: Low-rank adaptation of large language models.
\newblock In \emph{International Conference on Learning Representations}, 2022.
\newblock URL \url{https://openreview.net/forum?id=nZeVKeeFYf9}.

\bibitem[Huang et~al.(2021)Huang, Ma, Erfani, Bailey, and Wang]{huang2021unlearnable}
Huang, H., Ma, X., Erfani, S.~M., Bailey, J., and Wang, Y.
\newblock Unlearnable examples: Making personal data unexploitable.
\newblock In \emph{International Conference on Learning Representations}, 2021.
\newblock URL \url{https://openreview.net/forum?id=iAmZUo0DxC0}.

\bibitem[Huang et~al.(2023)Huang, Jiang, Dong, Qiao, Gao, and Li]{huang2023instruct2act}
Huang, S., Jiang, Z., Dong, H., Qiao, Y., Gao, P., and Li, H.
\newblock Instruct2act: Mapping multi-modality instructions to robotic actions with large language model.
\newblock \emph{arXiv preprint arXiv:2305.11176}, 2023.

\bibitem[Li et~al.(2018)Li, Farkhoor, Liu, and Yosinski]{li2018measuring}
Li, C., Farkhoor, H., Liu, R., and Yosinski, J.
\newblock Measuring the intrinsic dimension of objective landscapes.
\newblock 2018.

\bibitem[Li et~al.(2023)Li, Liu, and Gao]{li2023make}
Li, X., Liu, M., and Gao, S.
\newblock Make text unlearnable: Exploiting effective patterns to protect personal data, 2023.

\bibitem[Li \& Liang(2021)Li and Liang]{li-liang-2021-prefix}
Li, X.~L. and Liang, P.
\newblock Prefix-tuning: Optimizing continuous prompts for generation.
\newblock In \emph{Proceedings of the 59th Annual Meeting of the Association for Computational Linguistics and the 11th International Joint Conference on Natural Language Processing (Volume 1: Long Papers)}, pp.\  4582--4597, Online, August 2021. Association for Computational Linguistics.
\newblock \doi{10.18653/v1/2021.acl-long.353}.
\newblock URL \url{https://aclanthology.org/2021.acl-long.353}.

\bibitem[Loshchilov \& Hutter(2019)Loshchilov and Hutter]{loshchilov2019decoupled}
Loshchilov, I. and Hutter, F.
\newblock Decoupled weight decay regularization, 2019.

\bibitem[Luo et~al.(2023)Luo, Zhang, Fan, Yang, Wu, Qiao, and Nie]{luo2023biomedgpt}
Luo, Y., Zhang, J., Fan, S., Yang, K., Wu, Y., Qiao, M., and Nie, Z.
\newblock Biomedgpt: Open multimodal generative pre-trained transformer for biomedicine.
\newblock \emph{arXiv preprint arXiv:2308.09442}, 2023.

\bibitem[Nguyen et~al.(2022)Nguyen, Huynh, Nguyen, Liew, Yin, and Nguyen]{nguyen2022survey}
Nguyen, T.~T., Huynh, T.~T., Nguyen, P.~L., Liew, A. W.-C., Yin, H., and Nguyen, Q. V.~H.
\newblock A survey of machine unlearning, 2022.

\bibitem[OpenAI(2022)]{chatgpt}
OpenAI.
\newblock {Introducing ChatGPT}.
\newblock \url{https://openai.com/blog/chatgpt}, 2022.

\bibitem[Ouyang et~al.(2022)Ouyang, Wu, Jiang, Almeida, Wainwright, Mishkin, Zhang, Agarwal, Slama, Ray, et~al.]{ouyang2022training}
Ouyang, L., Wu, J., Jiang, X., Almeida, D., Wainwright, C., Mishkin, P., Zhang, C., Agarwal, S., Slama, K., Ray, A., et~al.
\newblock Training language models to follow instructions with human feedback.
\newblock \emph{Advances in Neural Information Processing Systems}, 35:\penalty0 27730--27744, 2022.

\bibitem[Peng et~al.(2023{\natexlab{a}})Peng, Wu, Allard, Kilpatrick, and Heidel]{openaiFinetune}
Peng, A., Wu, M., Allard, J., Kilpatrick, L., and Heidel, S.
\newblock Gpt-3.5 turbo fine-tuning and api updates, 8 2023{\natexlab{a}}.
\newblock URL \url{https://openai.com/blog/gpt-3-5-turbo-fine-tuning-and-api-updates}.

\bibitem[Peng et~al.(2023{\natexlab{b}})Peng, Wu, Allard, Kilpatrick, and Heidel]{peng2023gpt35}
Peng, A., Wu, M., Allard, J., Kilpatrick, L., and Heidel, S.
\newblock Gpt-3.5 turbo fine-tuning and api updates, August 2023{\natexlab{b}}.
\newblock URL \url{https://openai.com/blog/gpt-3-5-turbo-fine-tuning-and-api-updates}.
\newblock Illustration: Ruby Chen.

\bibitem[Qi et~al.(2023)Qi, Zeng, Xie, Chen, Jia, Mittal, and Henderson]{qi2023finetuning}
Qi, X., Zeng, Y., Xie, T., Chen, P.-Y., Jia, R., Mittal, P., and Henderson, P.
\newblock Fine-tuning aligned language models compromises safety, even when users do not intend to!, 2023.

\bibitem[Ren et~al.(2022)Ren, Xu, Wan, Ma, Sun, and Tang]{ren2022transferable}
Ren, J., Xu, H., Wan, Y., Ma, X., Sun, L., and Tang, J.
\newblock Transferable unlearnable examples.
\newblock \emph{arXiv preprint arXiv:2210.10114}, 2022.

\bibitem[Rozière et~al.(2023)Rozière, Gehring, Gloeckle, Sootla, Gat, Tan, Adi, Liu, Remez, Rapin, Kozhevnikov, Evtimov, Bitton, Bhatt, Ferrer, Grattafiori, Xiong, Défossez, Copet, Azhar, Touvron, Martin, Usunier, Scialom, and Synnaeve]{codeLlama}
Rozière, B., Gehring, J., Gloeckle, F., Sootla, S., Gat, I., Tan, X.~E., Adi, Y., Liu, J., Remez, T., Rapin, J., Kozhevnikov, A., Evtimov, I., Bitton, J., Bhatt, M., Ferrer, C.~C., Grattafiori, A., Xiong, W., Défossez, A., Copet, J., Azhar, F., Touvron, H., Martin, L., Usunier, N., Scialom, T., and Synnaeve, G.
\newblock Code llama: Open foundation models for code, 8 2023.
\newblock URL \url{https://ai.meta.com/research/publications/code-llama-open-foundation-models-for-code/}.

\bibitem[Sekhari et~al.(2021)Sekhari, Acharya, Kamath, and Suresh]{NEURIPS2021_9627c45d}
Sekhari, A., Acharya, J., Kamath, G., and Suresh, A.~T.
\newblock Remember what you want to forget: Algorithms for machine unlearning.
\newblock In Ranzato, M., Beygelzimer, A., Dauphin, Y., Liang, P., and Vaughan, J.~W. (eds.), \emph{Advances in Neural Information Processing Systems}, volume~34, pp.\  18075--18086. Curran Associates, Inc., 2021.
\newblock URL \url{https://proceedings.neurips.cc/paper_files/paper/2021/file/9627c45df543c816a3ddf2d8ea686a99-Paper.pdf}.

\bibitem[Shazeer \& Stern(2018)Shazeer and Stern]{shazeer2018adafactor}
Shazeer, N. and Stern, M.
\newblock Adafactor: Adaptive learning rates with sublinear memory cost, 2018.

\bibitem[Touvron et~al.(2023)Touvron, Lavril, Izacard, Martinet, Lachaux, Lacroix, Rozière, Goyal, Hambro, Azhar, Rodriguez, Joulin, Grave, and Lample]{touvron2023llama}
Touvron, H., Lavril, T., Izacard, G., Martinet, X., Lachaux, M.-A., Lacroix, T., Rozière, B., Goyal, N., Hambro, E., Azhar, F., Rodriguez, A., Joulin, A., Grave, E., and Lample, G.
\newblock Llama: Open and efficient foundation language models, 2023.

\bibitem[Wang et~al.(2023)Wang, Kordi, Mishra, Liu, Smith, Khashabi, and Hajishirzi]{wang2023selfinstruct}
Wang, Y., Kordi, Y., Mishra, S., Liu, A., Smith, N.~A., Khashabi, D., and Hajishirzi, H.
\newblock Self-instruct: Aligning language models with self-generated instructions, 2023.

\bibitem[Yang et~al.(2023)Yang, Wang, Zhang, Petzold, Wang, Zhao, and Lin]{yang2023shadow}
Yang, X., Wang, X., Zhang, Q., Petzold, L., Wang, W.~Y., Zhao, X., and Lin, D.
\newblock Shadow alignment: The ease of subverting safely-aligned language models, 2023.

\bibitem[Zhang et~al.(2023)Zhang, Ma, Yi, Sang, Jiang, Wang, and Xu]{zhang2023unlearnable}
Zhang, J., Ma, X., Yi, Q., Sang, J., Jiang, Y.-G., Wang, Y., and Xu, C.
\newblock Unlearnable clusters: Towards label-agnostic unlearnable examples.
\newblock In \emph{Proceedings of the IEEE/CVF Conference on Computer Vision and Pattern Recognition}, pp.\  3984--3993, 2023.

\bibitem[Zhou et~al.(2023)Zhou, Liu, Xu, Iyer, Sun, Mao, Ma, Efrat, Yu, Yu, Zhang, Ghosh, Lewis, Zettlemoyer, and Levy]{zhou2023lima}
Zhou, C., Liu, P., Xu, P., Iyer, S., Sun, J., Mao, Y., Ma, X., Efrat, A., Yu, P., Yu, L., Zhang, S., Ghosh, G., Lewis, M., Zettlemoyer, L., and Levy, O.
\newblock Lima: Less is more for alignment, 2023.

\bibitem[Zhu \& Li(2023)Zhu and Li]{zhu2023physics}
Zhu, Z.~A. and Li, Y.
\newblock Physics of language models: Part 3.1, knowledge storage and extraction, 2023.

\end{thebibliography}
\bibliographystyle{icml2023}

\newpage
\appendix
\onecolumn
\section{Appendix}
We will add the appendix in the future version.

\end{document}